\documentclass[twoside,11pt]{article}
 
 \usepackage[abbrvbib, nohyperref, preprint]{jmlr2e}

 \usepackage[table, svgnames, dvipsnames]{xcolor}
 \usepackage[utf8]{inputenc} 
 \usepackage{hyperref}       
 \usepackage{natbib}
 \usepackage{amsmath}
 \usepackage{mathtools}
 \usepackage{amssymb}
 \usepackage{subfigure}
 \usepackage{booktabs}
 \usepackage{todo}
 \usepackage[capitalise,nameinlink]{cleveref}
 
 \usepackage[normalem]{ulem}

\usepackage{thmtools}
\declaretheorem[
mdframed={
  skipabove=6pt,
  skipbelow=6pt,
  hidealllines=true,
  backgroundcolor={lightgray},
  innerleftmargin=8pt,
  innerrightmargin=8pt}
]{ex}

\newcommand{\grad}{\nabla}

\newcommand{\vnatparam}{\vlambda}

\newcommand{\vmeanparam}{\vmu}

\newcommand{\dkls}[3]{\mathbb{D}_{\text{KL}}^{#1}[#2 \, \|\, #3]}

\newcommand\cut[1]{}

\newcommand{\squishlist}{
   \begin{list}{$\bullet$}
    { \setlength{\itemsep}{0pt}      \setlength{\parsep}{3pt}
      \setlength{\topsep}{3pt}       \setlength{\partopsep}{0pt}
      \setlength{\leftmargin}{1.5em} \setlength{\labelwidth}{1em}
      \setlength{\labelsep}{0.5em} } }

\newcommand{\squishlisttwo}{
   \begin{list}{$\bullet$}
    { \setlength{\itemsep}{0pt}    \setlength{\parsep}{0pt}
      \setlength{\topsep}{0pt}     \setlength{\partopsep}{0pt}
      \setlength{\leftmargin}{2em} \setlength{\labelwidth}{1.5em}
      \setlength{\labelsep}{0.5em} } }

\newcommand{\squishend}{
    \end{list}  }

{}
\newtheorem{thm}{Theorem}{}
{}
{}

\newcommand{\half}{\mbox{$\frac{1}{2}$}}

\newcommand{\rnd}[1]{\left(#1\right)}
\newcommand{\sqr}[1]{\left[#1\right]}

\newcommand{\myang}[1]{\langle#1\rangle}
\newcommand{\myexpect}{\mathbb{E}}

\newcommand{\gauss}{\mbox{${\cal N}$}}

\newcommand{\myvec}[1]{\mbox{$\mathbf{#1}$}}
\newcommand{\myvecsym}[1]{\mbox{$\boldsymbol{#1}$}}

\newcommand{\veta}{\mbox{$\myvecsym{\eta}$}}

\newcommand{\vmu}{\mbox{$\myvecsym{\mu}$}}

\newcommand{\vlambda}{\mbox{$\myvecsym{\lambda}$}}

\newcommand{\vm}{\mbox{$\myvec{m}$}}

\newcommand{\vu}{\mbox{$\myvec{u}$}}
\newcommand{\vv}{\mbox{$\myvec{v}$}}

\newcommand{\vy}{\mbox{$\myvec{y}$}}

\newcommand{\vz}{\mbox{$\myvec{z}$}}

\newcommand{\vI}{\mbox{$\myvec{I}$}}

\newcommand{\vS}{\mbox{$\myvec{S}$}}
\newcommand{\vT}{\mbox{$\myvec{T}$}}
\newcommand{\vU}{\mbox{$\myvec{U}$}}
\newcommand{\vV}{\mbox{$\myvec{V}$}}
\newcommand{\vW}{\mbox{$\myvec{W}$}}

\newcommand{\vY}{\mbox{$\myvec{Y}$}}
\newcommand{\vZ}{\mbox{$\myvec{Z}$}}

\newcommand{\trace}{\mbox{Tr}}

\newcommand{\be}{\begin{equation}}
\newcommand{\ee}{\end{equation}}
\newcommand{\bea}{\begin{eqnarray}}
\newcommand{\eea}{\end{eqnarray}}
\newcommand{\beaa}{\begin{eqnarray*}}
\newcommand{\eeaa}{\end{eqnarray*}}

\crefname{section}{Sec.}{Sections}
\crefname{appendix}{App.}{Appendices}
\crefname{algorithm}{Alg.}{Algorithms}
\crefname{equation}{Eq.}{Eqs.}
\crefname{figure}{Fig.}{Figures}
\creflabelformat{equation}{#2\textup{#1}#3} 

 \ShortHeadings{VB Made Easy}{Khan}
 \firstpageno{1}

 \title{Variational Bayes Made Easy}

 \author{\name Mohammad Emtiyaz Khan \email emtiyaz.khan@riken.jp \\
        \addr RIKEN Center for Advanced Intelligence Project\\
        1-4-1 Nihonbashi, Chuo-ku, Tokyo 103-0027, Japan
        }

 \editor{}

\begin{document}

\maketitle

\begin{abstract}
   Variational Bayes is a popular method for approximate inference but its derivation can be cumbersome. To simplify the process, we give a 3-step recipe to identify the posterior form by explicitly looking for linearity with respect to expectations of well-known distributions. We can then directly write the update by simply ``reading-off'' the terms in front of those expectations. The recipe makes the derivation easier, faster, shorter, and more general. 
\end{abstract}


\section{Introduction}

Since its introduction in the early 90s \citep{hinton1993keeping,saul1996mean,Jaakkola96b}, variational Bayes (VB) has become a prominent method for approximate inference but, despite significant progress in probabilistic programming, deriving VB updates remains cumbersome for many. The main source of difficulty is the need to derive closed-form expressions for the integrals. For example, consider observed data $\vy$ and latent vector $\vz = (\vz_1,\vz_2,\ldots,\vz_K)$ for a model $p(\vy,\vz)$, 
and assume that we seek a mean-field posterior approximation:
   $p(\vz|\vy) \approx q(\vz) = \prod_{i} q_i(\vz_i)$.
Then, a stationary point $q^*(\vz)$ of the evidence lower-bound (ELBO) satisfies the following \citep[Eq. 10.9]{bishop2006pattern}, 
\begin{align}
   \log q_i^*(\vz_i) = \myexpect_{q^*_{\backslash i}} \sqr{ \log p(\vy,\vz) }  + \text{const.},
   \label{eq:vb_fixed_pt}
\end{align}
where the integral in the right hand side marginalizes out the rest of the variables (denoted by $\vz_{\backslash i}$) using the marginal $q^*_{\backslash i}(\vz_{\backslash i})$; see a proof in \cref{app:linearity}.

For conjugate-exponential (CE) family models, the integrals have closed-form expressions, but deriving them can still be cumbersome. It is not uncommon to find papers where these derivations take many pages of work. 
With such long derivations, it is easy to lose the main intuition and elegance of the updates. Difficulty increases when non-conjugate factors are also present. For such cases, despite using automatic differentiation, we still need to make several difficult choices: Which optimizer to choose and are they compatible with the conjugate updating? How to set the learning rate so that overall updates converge? And, when using coordinate-wise updates, in what
sequence should we update the variables? Here, we present a recipe to derive the updates that simplify these difficulties.

Our recipe consists of 3-steps based on an alternate way to express \cref{eq:vb_fixed_pt}. We assume an exponential-family form for 
   $q_{\text{\vlambda}_i}(\vz_i) \propto \exp\rnd{\myang{\text{\vT}_i(\text{\vz}_i), \text{\vnatparam}_i}}$
with natural parameter $\vlambda_i$, sufficient statistics $\vT_i(\cdot)$, and a constant base-measure. We can then write \cref{eq:vb_fixed_pt} in a form shown below that uses the pair $(\vlambda_i, \vmu_i)$ where $\vmu_i(\vnatparam_i) = \myexpect_{q_{\text{\vnatparam}_i}}[\vT_i(\vz_i)]$ is the expectation parameter: 
\begin{equation}
   \vlambda_i^* = \nabla_{\text{\vmu}_i} \left. \myexpect_{q_{\text{\vnatparam}}} \sqr{ \log p(\vy,\vz) } \right\vert_{\text{\vmu}_i = \text{\vmu}_i^*}.
   \label{eq:blr_fixed_pt}
\end{equation}
Here, we marginalize with respect to $q_{\text{\vnatparam}}(\vz)$ where $\vlambda = (\vlambda_1,\vlambda_2,\ldots, \vlambda_K)$ and denote by $(\vlambda_i^*, \vmu_i^*)$ the natural and expectation parameter pair for the optimal $q_i^*(\vz_i)$.
A proof is given in \cref{app:linearity} by using results from \citet{khan2017conjugate}, along with an extension for non-constant base-measures. The advantage of this expression is that we do not have to think about the distributional forms of $q_i$ or any integrals. Rather, we can simply look up the definition of $(\vlambda_i, \vmu_i)$; see \cref{tab:EFsummary} for a list.

We give an example below. For notational ease, we will write $\myexpect_{q_\text{\vnatparam}}[\cdot]$ as $\myexpect_{q}[\cdot]$.
   \begin{ex}
      \label{ex:mm}
      {\bf (A simple mixture model)} We consider two components $p_a(\vy)$ and $p_b(\vy)$ for an observation $\vy$ where mixture indicator $z\in\{0,1\}$ is sampled from a Bernoulli prior,
   \begin{align*}
      p(\vy|z) = p_a(\vy)^z p_b(\vy)^{1-z}, \text{ where } p(z) = \pi_0^z (1-\pi_0)^{1-z}.
   \end{align*}
   The posterior over $z$ can be obtained by using Bayes' rule,
      $p(z=1|\vy) = \frac{\pi_0 p_a(\text{\vy})}{\pi_0 p_a(\text{\vy}) + (1-\pi_0) p_b(\text{\vy})}$, but let us suppose that we do not know the posterior and we want to recover it from \cref{eq:blr_fixed_pt}.

      Denoting the posterior by $q(z)$, we can figure out its form by expanding  
   \begin{equation}
      \begin{split}
         \myexpect_{q}[\log p(\vy,z)] &= \myexpect_q[\log p(\vy|z) + \log p(z)]\\
         &=\myexpect_{q} \sqr{ z \log p_a(\vy) + (1-z) \log p_b(\vy) + z \log \pi_0 + (1-z) \log (1-\pi_0) } \\
         &= \underbrace{ \myexpect_{q}(z) }_{=\mu} \underbrace{ \log \frac{ \pi_0 p_a(\vy) }{ (1-\pi_0) p_b(\vy)} }_{\text{Coeff. in front of $\mu$}} +  \underbrace{\log [(1-\pi_0)p_b(\vy)]}_{\text{constant}} . 
      \end{split}
      \label{eq:Eloss_for_simpleMM}
   \end{equation}
      The last line suggests to choose $q(z)$ with expectation parameter $\mu = \myexpect_{q}(z)$. From \cref{tab:EFsummary}, we find Bernoulli distribution to have this expectation parameter, therefore we set $q(z) \propto \pi_1^z (1-\pi_1^{1-z})$ where $\pi_1>0$ is the probability of $z=1$ under $q$.

      Having figured out the form of $q$, we can now use \cref{eq:blr_fixed_pt} to find its parameter. For this we first look up from \cref{tab:EFsummary} the natural parameter: $\lambda = \log \frac{\pi_1}{1-\pi_1}$ and set it according to \cref{eq:blr_fixed_pt}.
      It is clear from \cref{eq:Eloss_for_simpleMM} that $\myexpect_{q} \sqr{ \log p(\vy,z) }$ is linear in $\mu$, therefore the gradient $\nabla_{\mu} \myexpect_{q} \sqr{ \log p(\vy,z) }$ is simply the coefficient in front of $\mu$. This gives us 
   \begin{equation}
      \lambda^* = \log \frac{ \pi_0 p_a(\vy) }{ (1-\pi_0) p_b(\vy)} 
      \quad \implies \pi_{1}^* = \frac{\pi_0 p_a(\vy)}{\pi_0 p_a(\vy) + (1-\pi_0) p_b(\vy)}.
      \label{eq:mm_finalupdate}
   \end{equation}
      where in the second equality we rewrite the update in terms of $\pi_1$.
      This recovers the posterior $p(z=1|\vy)$ obtained with the Bayes' rule.
   \end{ex}
The procedure is an instance of our 3-step recipe shown below where we identify the posterior form by explicitly looking for linearity with respect to $\vmu$ and use it to update $\vlambda$:
\begin{enumerate}
   \item Identify all $q_i$ at once, by looking for linearity of $\myexpect_q[\log p(\vy,\vz)]$ w.r.t. some $\vmu_i$.
   \item Compute $\nabla_{\text{\vmeanparam}_i} \myexpect_q[\log p(\vy,\vz)]$ (hint: for CE, just read-off the coefficient in front of $\vmeanparam_i$).
   \item Update $\vnatparam_i \leftarrow (1-\rho_i)\vnatparam_i + \rho_i \grad_{\text{\vmeanparam}_i} \myexpect_q[\log p(\vy,\vz)]$ with a learning rate $\rho_i$ (often set to 1).
\end{enumerate}
For most cases, there is no need to derive any integrals: we just expand $\myexpect_q[\log p(\vy,\vz)]$ as a multi-linear function of all $\vmu_i$ at once and look-up the rest from a table (similar to \cref{tab:EFsummary}).
We can then directly write the update by simply ``reading-off'' the coefficient in front of $\vmu_i$, which is now a function of the $\vmu_{\backslash i}$ and observation $\vy$.

The steps differ fundamentally from the standard way of deriving VB where each $q_i$ is identified \emph{separately}. Such derivations can be very long and tedious because we need to investigate each node separately by looking for a known distributional form for each of them~\citep[App. A]{blei2017variational}\citep[Sec. 10.2]{bishop2006pattern}. 
The distributional form is not used anyways because the updates are ultimately written and implemented using the parameters of the distributions.
Our recipe makes the direct use of the $(\vlambda_i,\vmu_i)$ pair, and leads to an easier, faster, and shorter derivation where all nodes are handled altogether.

The recipe is also more general and covers a wide-variety of cases, including conjugate, non-conjugate, and even deterministic factors.
This is due to Step 3 which uses the Bayesian learning rule (BLR) \citep{khan2021bayesian} that contains many algorithms as special cases. Below, we give additional sub-steps for Step 3 to derive many variants:
      \begin{enumerate}
         \item[(3a)] Derive Bayes-rule by using $\rho_i = 1$; rewrite it in your parameter of choice (see \cref{ex:mm}).
         \item[(3b)] Derive Coordinate Ascent VI (CAVI) and variational message passing (VMP) by setting $\rho_i=1$ and doing a coordinate-wise update (see \cref{ex:mm2} and \cref{ex:gmm}).
         \item[(3c)] Derive Stochastic VI (SVI) by setting $\rho_i=1$ for the local variables and updating the global variable after a local update on any node $i$ (see \cref{ex:mm_svi}).
         \item[(3d)] Derive updates of deterministic nodes (and Laplace's approximation) by invoking the delta method to approximate the expectation (see \cref{ex:mf}).
         \item[(3e)] In presence of the non-conjugate terms, just use the gradient and simplify it using reparameterization trick whenever feasible \citep{khan2021bayesian} (see \cref{ex:nc}).
      \end{enumerate}
Another benefit is that, Steps 3a-3d do not only not require any integrals but we do not even need any derivatives either; due to linearity we just need the coefficients in front. With Step 3d, \emph{even MAP estimates can be derived without any derivatives}.
Moreover, there is no need to use different optimizers for conjugate and non-conjugate parts, and the learning rate and update schedule need not be set carefully because the
update converges under fairly general conditions \citep{khan2016faster}. In contrast, coordinate-wise update may not always converge and require the solution to be unique for each coordinate \citep{paquetconvergence}.

\begin{table}[!t]
   \begin{center}
      \caption{Exponential-family used in this paper. For Gaussian-Wishart, we use two sets of variables: $\vz_1$ is a real vector and $\vZ_2$ is a positive-definite matrix. We do not give the expressions for the expectation parameters explicitly but these can be found in \href{https://en.wikipedia.org/wiki/Exponential_family\#Table_of_distributions}{Wikipedia}, with a more exhaustive list covering many other distributions.\vspace{.05in}}
   \label{tab:EFsummary}
         {\def\arraystretch{1.2}
         \begin{tabular}{p{0.6in} p{1.5in} l l}
         \toprule
            Name & Distribution $q(\vz)$ & Expectation Param $\vmu$ & Natural param $\vlambda$ \\
         \bottomrule
         \rowcolor{Gainsboro!60}
            Bernoulli & $ \propto\pi^z(1-\pi^{1-z})$ & $\myexpect_q(z)$ & $\log \frac{\pi}{1-\pi}$ \\
          Beta & $ \propto z^{\alpha-1}e^{-\beta z}$ & $\myexpect_q \sqr{ \begin{array}{l} \log z \\ \log(1-z)\end{array} }$ & $\sqr{ \begin{array}{l} \alpha - 1\\ \beta-1 \end{array} }$ \\
         \rowcolor{Gainsboro!60}
             Gaussian  & $ \propto e^{-\frac{1}{2}(\text{\vz}-\text{\vm})^\top\text{\vS} (\text{\vz}-\text{\vm})} $ & $\myexpect_q \sqr{ \begin{array}{l} \vz\\ \vz\vz^\top \end{array} }$ & $\sqr{ \begin{array}{l} \vS\vm \\ -\half\vS \end{array} }$ \\
                Gaussian-Wishart & $\propto |\vZ_2|^{\frac{1}{2}} e^{-\frac{1}{2}(\text{\vz}_1-\text{\vm})^\top\gamma\text{\vZ}_2 (\text{\vz}_1-\text{\vm})}$ $|\vZ_2|^{\frac{\nu-D-1}{2}}e^{-\frac{1}{2} \text{\trace}{(\text{\vW}^{-1} \text{\vZ}_2 )}} $ & $\myexpect_q \sqr{ \begin{array}{l} \log|\vZ_2| \\ \vZ_2 \\ \vZ_2\vz_1 \\ \vz_1^\top \vZ_2\vz_1 \end{array}}$ & $\sqr{ \begin{array}{l} \half(\nu-D) \\ -\half(\vW^{-1} + \gamma\vm\vm^\top) \\ \gamma\vm \\ -\half \gamma\end{array}}$ \\
         \bottomrule
      \end{tabular}
   }
   \end{center}
\end{table}

\section{Examples}
Let us see another example with a slightly more complex model where a mean-field approximation is required. We will see that, unlike standard derivation where each factor requires a separate expansion of the log-joint, we can identify both distributions simultaneously. Another advantage is that we do not have to start with a mean-field structure, rather we can figure this out by simply looking for linearity.

\begin{ex}
   \label{ex:mm2}
   {\bf (A mixture-model with two levels)}
   We will add an additional latent variable: $\pi_0 \rightarrow z_i \rightarrow \vy_i$ where we have multiple \emph{vector} observations $\vy_i$, and two sets of latent variables: $z_i$ and $\pi_0$. We assume the following model with a beta prior over $\pi_0$,
   \begin{equation*}
   \begin{split}
      p(\vY|\vz) &= \prod_{i=1}^N p_a(\vy_i)^{z_i} p_b(\vy_i)^{1-z_i},\,\,
      p(z_i|\pi_0) = \pi_0^{z_i} (1-\pi_0)^{1-z_i},\,\,
      p(\pi_0) \propto \pi_0^{\alpha_0-1} (1-\pi_0)^{\beta_0 -1}
   \end{split}
   \end{equation*}
   Our goal is to estimate $p(\vz, \pi_0|\vY)$ but the model is not a conjugate one, therefore we have to use a mean-field approximation for tractability.

   In the $1^\text{st}$ step we can figure out $q$ for both $z_i$ and $\pi_0$ at once by using linearity in the expected log-joint; there is no need to decide the factorization yet. We show this below where the second-line is obtained by using \cref{eq:Eloss_for_simpleMM},
\begin{align}
   &\myexpect_q[\log p(\vY|\vz)p(\vz) p(\pi_0)] = \myexpect_q\sqr{
      \sum_{i=1}^N \rnd{\log p(\vy_i|z_i) + \log p(z_i) } +  \log p(\pi_0)} \nonumber\\
   &= \myexpect_q\sqr{ \sum_{i=1}^N \rnd{ z_i \log \frac{ \pi_0 p_a(\vy_i) }{ (1-\pi_0) p_b(\vy_i) } + \log [(1-\pi_0) p_b(\vy_i)] } +  \log \rnd{ \pi_0^{(\alpha_0-1)}(1-\pi_0)^{(\beta_0-1) } } } \nonumber\\
   &= \sum_{i=1}^N \underbrace{ \myexpect_q(z_i) }_{=\mu_i} \Bigg( \underbrace{ \myexpect_q{\sqr{ \begin{array}{c} \log \pi_0\\  \log (1-\pi_0) \end{array} }}}_{=\text{\vmu}_0} + \sqr{ \begin{array}{c} \log p_a(\vy_i)\\  \log p_b(\vy_i) \end{array} }  \Bigg)^\top \sqr{ \begin{array}{c} +1\\  -1 \end{array} }  + \sum_i \log p_b(\vy_i)\nonumber\\
      &\qquad\qquad \qquad \qquad \qquad \qquad \qquad  +  { \underbrace{ \myexpect_q{\sqr{ \begin{array}{c} \log \pi_0\\  \log (1-\pi_0) \end{array} }}}_{=\text{\vmu}_0} }^\top\sqr{ \begin{array}{c} \alpha_0-1 \\ N+\beta_0-1 \end{array} } ,
         \label{eq:mm2_expand}
\end{align}
   where in the last step we assumed that $q(\vz, \pi_0) = \prod_i q(z_i) q(\pi_0)$ to get linearity with respect to both $\mu_1$ and $\vmu_0$. Using \cref{tab:EFsummary}, we have $q(z_i)$ as Bernoulli and $q(\pi_0)$ as Beta. 

   After this, the $2^\text{nd}$ step is straightforward to simply read the coefficient in front of $\mu_i$ and $\vmu_0$ respectively, and use them in the $3^\text{rd}$ step with $\rho_i =1$. This is shown below in both natural parameters
   and posterior parameters of Bernoulli (denoted by $\pi_{i,1}$) and Beta distributions (denoted by $(\alpha_1, \beta_1)$) obtained by using \cref{tab:EFsummary}:
   \begin{align}
      \lambda_i &\leftarrow \Bigg( \vmu_0 + \sqr{ \begin{array}{c} \log p_a(\vy_i)\\  \log p_b(\vy_i) \end{array} }  \Bigg)^\top \sqr{ \begin{array}{c} 1\\  -1 \end{array} } 
         \implies \pi_{i,1} \leftarrow \frac{e^{\myexpect_q[\log \pi_0]} p_a(\vy)}{e^{\myexpect_q[\log \pi_0]} p_a(\vy) + e^{\myexpect_q[\log (1-\pi_0)]} p_b(\vy)} \nonumber\\
         \vlambda_0 &\leftarrow  \sum_{i=1}^N \mu_i \sqr{ \begin{array}{c} +1\\ -1 \end{array} } + \sqr{ \begin{array}{c} \alpha_0-1 \\ N+\beta_0-1 \end{array} } \implies 
            \sqr{ \begin{array}{c} \alpha_1 \\ \beta_1 \end{array} } \leftarrow  \sqr{ \begin{array}{c} \alpha_0 + \sum_i \myexpect_q(z_i) \\ \beta_0  + \sum_i (1- \myexpect_q(z_i)) \end{array} }
   \end{align}
\end{ex}
The recipe identifies the distributions for all nodes at once by looking for linearity; see the last line in \cref{eq:mm2_expand}. In addition, it reuses the already-known integrals by looking-up the required expectations. For instance, in the example above, we need to find 
$\myexpect_q(z_i)$, $\myexpect_q(\log \pi_0)$, and $\myexpect_q(\log (1-\pi_0))$, whose expressions can be found in Wikipedia pages for \href{https://en.wikipedia.org/wiki/Bernoulli_distribution}{Bernoulli} and \href{https://en.wikipedia.org/wiki/Beta_distribution}{Beta} distributions.

We are now ready to discuss the most-commonly used example for VB, which is the Gaussian mixture model. Unlike the standard derivation covered in \citep{bishop2006pattern}, we can derive all the updates with just a single expansion of the expected log-joint. The example clearly demonstrates the usefulness of the recipe in simplifying the derivation.

\begin{ex}
   \label{ex:gmm}
   {\bf (Gaussian mixture model)} The model uses a Gaussian likelihood for the two components and a Gaussian-Wishart prior on their means and covariances,
   \begin{align*}
      p_a(\vy_i) &= \gauss(\vy_i|\vm_a, \vS_a^{-1}), \quad\quad 
      &&p(\vm_a, \vS_a) = \gauss\rnd{\vm_a|0, (\gamma_0 \vS_a)^{-1}} \mathcal{W}(\vS_a|\vW_0, \nu_0) \\
      p_b(\vy_i) &= \gauss(\vy_i|\vm_b, \vS_b^{-1}), \quad\quad 
      &&p(\vm_b, \vS_b) = \gauss\rnd{\vm_b|0, (\gamma_0 \vS_b)^{-1}} \mathcal{W}(\vS_b|\vW_0, \nu_0) 
   \end{align*}
   for scalars $\gamma_0>0$ and $\nu_0>D-1$, and positive-definite matrix $\vW_0$. For this, we simply need to make two changes in \cref{eq:mm2_expand}. First, we replace $\log p_a(\vy_i)$ and $\log p_b(\vy_i)$ by their expected values. We show one of them below in terms of the required expectations,
   \begin{align*}
      &\myexpect_q[\log p_a(\vy_i)] = \half \myexpect_q[\log |\vS_a|] - \half \myexpect_q \sqr{ (\vy_i -\vm_a)^\top \vS_a (\vy_i-\vm_a) } + c\\
      &\quad\quad= \half \underbrace{ \myexpect_q[ \log |\vS_a| ] }_{\text{\vmu}_{p_a, 1}} 
      -\half \trace\big( \vy_i\vy_i^\top \underbrace{ \myexpect_q [\vS_a] }_{\text{\vmu}_{p_a, 2} } \big) 
      + \vy_i^\top \underbrace{\myexpect_q [\vS_a\vm_a] }_{\text{\vmu}_{p_a, 3} }
      - \half \underbrace{ \myexpect_q[ \vm_a^\top\vS_a\vm_a ] }_{\text{\vmu}_{p_a, 4} }  + c. 
   \end{align*}
   The term is linear in terms of the expectation-parameter $\vmu_{p_a,1:4}$ which corresponds to the Gaussian-Wishart distribution (\cref{tab:EFsummary}); see the Wikipedia pages for their expressions.
   Second, we need to add the contribution of the prior, which is also linear in $\vmu_{p_a,1:4}$,
   \begin{align*}
      &\myexpect_q[\log p(\vm_a, \vS_a)] = \half (\nu_0-D)\myexpect_q[\log |\vS_a|] -\half \trace\big( \vW_0^{-1} \myexpect_q[\vS_a] \big) -\half \myexpect_q[\vm_a^\top(\gamma_0\vS_a)\vm_a] + c \\
      &\quad\quad= \half (\nu_0-D) \vmu_{p_a,1} -\half \trace\big( \vW_0^{-1} \vmu_{p_a,3} \big) -\half \gamma_0 \vmu_{p_a,4} + c.
   \end{align*}
   Then, we expand the expected log-joint by using model definition in the first line, then by plugging \cref{eq:mm2_expand} to get the second line, and do rearrangement afterward to get linearity,
   \begin{flalign*}
      &\myexpect_q[\log p(\vY|\vz, \vm_{a:b}, \vS_{a:b})p(\vz) p(\pi_0) p(\vm_{a:b}, \vS_{a:b})] &&\\
      &= \myexpect_q[\log p(\vY|\vz, \vm_{a:b}, \vS_{a:b})p(\vz) p(\pi_0)] + \myexpect_q[\log p(\vm_{a:b}, \vS_{a:b})] &&
   \end{flalign*}
   \begin{align*}
      &=\sum_{i=1}^N \mu_i \Bigg( \vmu_0 + \sqr{ \begin{array}{c} \myexpect_q(\log p_a(\vy_i)) \\  \myexpect_q(\log p_b(\vy_i)) \end{array} }  \Bigg)^\top \sqr{ \begin{array}{c} +1\\  -1 \end{array} }  + \sum_i \myexpect_q[\log p_b(\vy_i)] +  \vmu_0^\top\sqr{ \begin{array}{c} \alpha_0-1 \\ N+\beta_0-1 \end{array} } \\
      &\qquad\qquad\qquad\qquad\qquad\qquad\qquad\qquad + \myexpect_q[\log p(\vm_a, \vS_a)] + \myexpect_q[\log p(\vm_b, \vS_b)]  \\
      &=\sum_{i=1}^N \mu_i \Bigg( \vmu_0 + 
          \sqr{\begin{array}{c} 
             \half \vmu_{p_a,1} -\half \trace\rnd{\vy_i\vy_i^\top \vmu_{p_a,2}} + \vy_i^\top \vmu_{p_a,3} -\half \vmu_{p_a,4}\\
             \half \vmu_{p_b,1} -\half \trace\rnd{\vy_i\vy_i^\top \vmu_{p_b,2}} + \vy_i^\top \vmu_{p_b,3}-\half \vmu_{p_a,4}
          \end{array} }
            \Bigg)^\top \sqr{ \begin{array}{c} +1\\  -1 \end{array} }  
            \\
               &\quad\quad\quad\quad+ \sum_i \rnd{ \half \vmu_{p_b,1} -\half \trace\big[ \vy_i \vy_i^\top \vmu_{p_b,2} \big] +  \vy_i^\top \vmu_{p_b,3} - \half \vmu_{p_b,4} } +  \vmu_0^\top\sqr{ \begin{array}{c} \alpha_0-1 \\ N+\beta_0-1 \end{array} } \\
      &\quad\quad\quad\quad+ \half (\nu_0-D)\vmu_{p_a,1} -\half \trace\rnd{\vW_0^{-1} \vmu_{p_a,2}} - \half \gamma_0 \vmu_{p_a,4}\\
      &\quad\quad\quad\quad+ \half (\nu_0-D)\vmu_{p_b,1} -\half \trace\rnd{\vW_0^{-1} \vmu_{p_b,2}} - \half \gamma_0 \vmu_{p_b,4}  + \text{const.}
   \end{align*}
   The last equation looks complicated but it is linear with respect to each $\mu_i, \vmu_0, \vmu_{p_a}, \vmu_{p_b}$, therefore we can simply look up the coefficient in front to write the update.
   The updates of $q(\pi_0)$ remain same as before because the coefficient is unchanged. For $q(z_i)$, the coefficient takes an expectation over $\log p_a$ and $ \log p_b$, giving us the following,
\begin{align}
      \pi_{i,1} &\leftarrow \frac{e^{\myexpect_q[\log \pi_0] + \myexpect_q[\log p_a(\text{\vy})]} }{e^{\myexpect_q[\log \pi_0]+ \myexpect_q[\log p_a(\text{\vy})]} + e^{\myexpect_q[\log (1-\pi_0)]+ \myexpect_q[\log p_b(\text{\vy})]} }
   \end{align}
   The update for the mean and covariance of the component is also obtained in a straightforward manner. For $q(\vm_a, \vS_a)$, for instance, we set its 4 natural parameters (given in \cref{tab:EFsummary}) to the coefficients in front of $\vmu_{p_a,1}$ to $\vmu_{p_a,4}$ respectively, and rearrange to get
   \begin{align}
      \sqr{ \begin{array}{l} \half(\nu_a-D) \\ -\half (\vW_a^{-1} + \gamma_a \vm_a\vm_a^\top) \\ \gamma_a \vm_a \\ -\half \gamma_a \end{array}} 
         &=\sqr{ \begin{array}{l} \half \sum_i \mu_i + \half (\nu_0 -D) \\ -\half \sum_i \mu_i \vy_i\vy_i^\top -\half \vW_o^{-1} \\ \sum_i \mu_i \vy_i \\ -\half \sum_i\mu_i -\half \gamma_0 \end{array}} \\
            \implies \sqr{ \begin{array}{l} \nu_a \\ \vW_a^{-1} \\ \vm_a \\ \gamma_a \end{array}}
               &=\sqr{ \begin{array}{l} \sum_i \mu_i + \nu_0 \\ \sum_i \mu_i \vy_i\vy_i^\top - \frac{(\sum_i \mu_i \text{\vy}_i)( \sum_i \mu_i \text{\vy}_i)^\top }{\sum_i\mu_i + \gamma_0} + \vW_0^{-1} \\ \frac{1}{\gamma_a}\sum_i \mu_i \vy_i \\ \sum_i\mu_i + \gamma_0 \end{array}} \nonumber
   \end{align}
   We encourage the reader to verify that these updates are same as ones derived in \citep[Eqs 10.49, 10.58, 10.60-10.63]{bishop2006pattern}, but unlike that derivation, all updates here are derived by using just one expansion of expected log-joint, which leads to a shorter and much faster derivation.
   \end{ex}

We now discuss Sub-steps (3b)-(3f) and show that, by using the BLR in Step 3, we can derive many other types of algorithms from the VB updates. It makes sense to start from a more Bayesian approach and make approximation when required, rather than doing it the other way. Using the BLR in Step 3 does exactly this. For example, as we saw in the previous examples, we can write a CAVI or VMP style update as special cases by choosing $\rho_i=1$, and doing coordinate-wise updates. But the same
approach can be used to derive SVI \citep{hoffman2013stochastic}, described below for completeness. 
\begin{ex}
   \label{ex:mm_svi}
   {\bf (SVI updates for the mixture model with two levels)}
   Instead of a coordinate-wise update that goes through all $z_i$, we can pick randomly just one $z_i$, do the update with $\rho_i = 1$ using \cref{eq:mm_finalupdate}, and after this we update the global variable $\pi_0$ with $\rho_0 <1$,
      \begin{align}
         \vlambda_0 &\leftarrow (1-\rho_0) \vlambda_0 + \rho_0 \sum_{i=1}^N \mu_i \sqr{ \begin{array}{c} +1\\ -1 \end{array} } + \sqr{ \begin{array}{c} \alpha_0-1 \\ N+\beta_0-1 \end{array} } \nonumber \\
            &\implies 
            \sqr{ \begin{array}{c} \alpha_1 \\ \beta_1 \end{array} } \leftarrow (1-\rho_0) \sqr{ \begin{array}{c} \alpha_1 \\ \beta_1 \end{array} } + \rho_0 \sqr{ \begin{array}{c} \alpha_0 + \sum_i \myexpect_q(z_i) \\ \beta_0 + N - \sum_i \myexpect_q(z_i) \end{array} }
   \end{align}
   Updating $\pi_0$ after every $z_i$ update can speed up convergence \citep{hoffman2013stochastic}.
\end{ex}
Now, we give another example on matrix factorization where we show how to derive updates for deterministic nodes and also for expectation maximization. We derive the well-known alternating least-squares and probabilistic PCA \citep{tipping1999probabilistic}. Both of these are obtained by using the delta method where the expectation parameter is approximated by using its mean: 
   $\myexpect_q(\vz \vz^\top) \approx \vm\vm^\top$,
where $\vm$ is the mean of $q$. More details on such delta method is given in \citet{khan2021bayesian}.
\begin{ex}
   \label{ex:mf}
Matrix factorization and latent factor models aim to fit data matrix $\vY$ of size $N\times D$ by using two sets of factors $\vU$ and $\vV$ respectively of sizes $N\times K$ and $D\times K$ where $K$ is the number of factors. The likelihood and the prior are given by,
   \[
      p(\vY|\vU, \vV) 
      = \prod_{i=1}^N \prod_{j=1}^D \gauss(y_{ij}|\vu_i^\top\vv_j, \vI), \quad
      p(\vu_i) = \gauss(\vu_i|0, \vI/\delta_u), \quad
      p(\vv_j) = \gauss(\vv_j|0, \vI/\delta_v)
   \]
   where $y_{ij}$ is the $ij$'th entry, $\vu_i$ is the $i$'th row of $\vU$, and $\vv_j$ is the $j$'th row of $\vV$.
   The most popular procedure is to use the alternating least-squares (ALS) procedure, but there is also expectation maximization \citep{tipping1999probabilistic} and VMP \cite{paquetconvergence}. These can all be derived by using our recipe for VB, as we show now.

   Following the $1^{\text{st}}$ step, we expand the expected log-joint and look for linearity,
   \begin{align*}
      &\myexpect_q[\log p(\vY,\vU, \vV)] =  \myexpect_q \sqr{ \sum_{i=1}^N \rnd{ \sum_{j=1}^D -\frac{1}{2} (y_{ij} - \vu_i^\top\vv_j)^2 - \frac{\delta_v}{2}\vv_j^\top\vv_j } - \frac{\delta_u}{2}\vu_i^\top\vu_i } + \textrm{const} \\
   \end{align*}
   \begin{align*}
      &= \sum_{i=1}^N \Bigg( \sum_{j=1}^D -\frac{1}{2} \Bigg( \trace \Bigg( \underbrace{ \myexpect_q(\vu_i\vu_i^\top) }_{= \text{\vmu}_{u_i}^{(2)}} \underbrace{ \myexpect_q( \vv_j \vv_j^\top)}_{= \text{\vmu}_{v_j}^{(2)}} \Bigg) - 2y_{ij} { \underbrace{ \myexpect_q(\vu_i)}_{= \text{\vmu}_{u_i}^{(1)}} }^\top \underbrace{ \myexpect_q(\vv_j) }_{\text{\vmu}_{v_j}^{(1)} } \Bigg)   \\
      & \qquad\qquad\qquad\qquad - \frac{\delta_v}{2}\trace \underbrace{ \myexpect_q\rnd{\vv_j\vv_j^\top} }_{= \text{\vmu}_{v_j}^{(2)}}  \Bigg) - \frac{\delta_u}{2}\trace \underbrace{ \myexpect_q\rnd{\vu_i\vu_i^\top} }_{= \text{\vmu}_{u_i}^{(2)}}  + \textrm{const}
   \end{align*}
   Here, we use mean-field approximation for all $\vu_i$ and $\vv_j$ because we want linearity with respect to all. The sufficient statistics correspond to Gaussian (third row in \cref{tab:EFsummary}).

   The $2^{\text{nd}}$ step is to simply read off as the coefficients in the front, and, by using the natural parameters given in \cref{tab:EFsummary}, we can directly write the update from the $3^{\text{rd}}$ step. The first two updates below use the natural parameters denoted by $(\vS_{u_i}\vm_{u_i}, -\vS_{u_i}/2)$ corresponding to $(\vmu_{u_i}^{(1)}, \vmu_{u_i}^{(2)})$, while the next two updates do the same for the variable $\vv_j$,
   \begin{align*}
      \vS_{u_i}\vm_{u_i}  &\leftarrow (1-\rho_i) \vS_{u_i}\vm_{u_i} + \rho_i { \sum_{j=1}^D  \vmu_{v_j}^{(1)}y_{ij}  } \\
      \vS_{u_i} &\leftarrow (1-\rho_i) \vS_{u_i} + \rho_i { \Bigg(\sum_{j=1}^D \vmu_{v_j}^{(2)} + \delta_u\vI_K  \Bigg)} \\
      \vS_{v_j}\vm_{v_j}  &\leftarrow (1-\rho_j) \vS_{v_j}\vm_{v_j} + \rho_j { \sum_{i=1}^N  \vmu_{u_i}^{(1) } y_{ij} } \\
      \vS_{v_j} &\leftarrow (1-\rho_j) \vS_{v_j} + \rho_j \Bigg( { \sum_{i=1}^N \vmu_{u_i}^{(2)} + \delta_v\vI_K }  \Bigg).
   \end{align*}
   We can then specialize these updates to derive ALS, EM, VB etc.
   \begin{enumerate}
      \item Setting $\rho_i = \rho_j =1$ and updating posteriors of $\vU$ and $\vV$ in alternate iterations (that is coordinate wise updates), we get the VMP update.
   \begin{equation}
   \begin{split}
      \vS_{u_i}\vm_{u_i}  &\leftarrow { \sum_{j=1}^D  \vmu_{v_j}^{(1)}y_{ij}  },
      \qquad\qquad \vS_{u_i} \leftarrow  \sum_{j=1}^D \vmu_{v_j}^{(2)} + \delta_u\vI_K \\
      \vS_{v_j}\vm_{v_j}  &\leftarrow { \sum_{i=1}^N  \vmu_{u_i}^{(1) } y_{ij} }, 
      \qquad\qquad \vS_{v_j} \leftarrow { \sum_{i=1}^N \vmu_{u_i}^{(2)} + \delta_v\vI_K } .
   \end{split}
   \label{eq:vb_mf}
   \end{equation}

      \item We can get the probabilistic PCA updates \citep{tipping1999probabilistic}, where we only compute the posterior wrt $\vU$ and assume $\vV$ to be deterministic. We can do this by making two more additional approximations,
         \begin{enumerate}
            \item denote $\vm_{v_j}$ by $\hat{\vv}_{j}$, and 
            \item use the delta approximation for $\vmu_{v_j}^{(2)} = \myexpect_{q}[\vv_j\vv_j^\top] \,\,\approx\,\, \myexpect_{q}[\vv_j] \myexpect_{q}[\vv_j]^\top = \hat{\vv}_{j} \hat{\vv}_{j}^\top$.
         \end{enumerate}
         With these, the update in \cref{eq:vb_mf} reduces to
   \begin{equation}
   \begin{split}
      \vS_{u_i} \vm_{u_i} &\leftarrow \sum_{j=1}^D \hat{\vv}_{j} y_{ij},\quad
      \vS_{u_i} \leftarrow \sum_{j=1}^D \hat{\vv}_{j} \hat{\vv}_{j}^\top + \delta_u\vI_K\\
      \hat{\vv}_{j}  &\leftarrow \rnd{ \sum_{i=1}^N \vmu_{u_i}^{(2)} + \delta_v\vI_K }^{-1} { \sum_{i=1}^N \vmu_{u_i}^{(1)} \, y_{ij}}.
      \label{eq:em}
   \end{split}
   \end{equation}
         These are equivalent to those derived in \citet[Eqs. 25-17]{bishop2006pattern}.

      \item Similarly, we get the ALS updates by further 
         \begin{enumerate} 
            \item denoting $\vm_{u_i}$ by $\hat{\vu}_{i}$, and
            \item adding the delta approximation:
            $\vmu_{u_i}^{(2)} = \myexpect_{q}[\vu_i\vu_i^\top] \,\,\approx\,\, \myexpect_{q}[\vu_i] \myexpect_{q}[\vu_i]^\top = \hat{\vu}_{i} \hat{\vu}_{i}^\top$. 
         \end{enumerate} 
         With these, \cref{eq:em} reduces to the following ALS scheme,
   \begin{equation}
      \begin{split}
      \hat{\vu}_{i}  &\leftarrow \Bigg( \sum_{j=1}^D \hat{\vv}_{j} \hat{\vv}_{j}^\top + \delta_u\vI_K \Bigg)^{-1} \sum_{j=1}^D \hat{\vv}_{j} y_{ij} , \\  
         \hat{\vv}_{j}  &\leftarrow \Bigg( \sum_{i=1}^N \hat{\vu}_{i} \hat{\vu}_{i}^\top + \delta_v\vI_K \Bigg)^{-1} \sum_{i=1}^N \hat{\vu}_{i}^\top y_{ij} , 
      \end{split}
      \label{eq:est_u}
   \end{equation}
   \end{enumerate}
   No derivatives are used to derive ALS (a MAP estimation procedure).
   It is also possible to update all quantities in parallel by using all learning rates $<1$, and convergence is less of an issue unlike VMP \citep{paquetconvergence}.
\end{ex}

Finally, we briefly discuss the inclusion of non-conjugate terms.
\begin{ex}
   \label{ex:nc}
   Suppose that, in \cref{ex:mm}, we decided to use a non-conjugate prior, say, a logit-normal distribution with mean $m$,
$   p(\pi_0) \propto \frac{1}{\pi_0 (1-\pi_0)} \exp\sqr{- \frac{(\text{logit}(\pi_0) - m)^2}{2}}.$
Still, the derivation proceeds in the same way by expanding $\myexpect_q[\log p(\vY|\vz)p(\vz) p(\pi_0)] =$ 
\begin{align*}
   \Bigg(\sum_{i=1}^N \underbrace{ \myexpect_q(z_i) }_{=\mu_i} \underbrace{ \myexpect_q{\sqr{ \begin{array}{c} \log \pi_0\\  \log (1-\pi_0) \end{array} }}^\top}_{=\text{\vmu}_0^\top} \underbrace{ \sqr{ \begin{array}{c} p_a(\vy_i)\\  p_b(\vy_i) \end{array} }}_{\text{Coeff. in front}}  \Bigg) + N \underbrace{ \myexpect_q[\log (1-\pi_0)] }_{\mu_0^{(2)}} + \underbrace{ \myexpect_q[\log p(\pi_0)]}_{\textrm{non-conjugate term}}.
\end{align*}
and combine the last two terms by collecting the linear terms together,
   \[
      \underbrace{ \myexpect_q{\sqr{ \begin{array}{c} \log \pi_0\\  \log (1-\pi_0) \end{array} }}^\top}_{=\text{\vmu}_0^\top} \underbrace{\sqr{ \begin{array}{c} -1 \\ N-1 \end{array} } }_{\text{Coeff. in front}} + \underbrace{ \myexpect_q\sqr{- \frac{(\text{logit}(\pi_0) - m)^2}{2}}}_{\textrm{non-conjugate term}}.
   \]
We will now just have to add the last term to the derivative of $\nabla_{\text{\vmu}_0}$, and rewrite the update. The form of the update does not change much, rather the non-conjugate prior can simply be written as a ``pseudo'' conjugate Beta prior,
   \[
      \vlambda_0 \leftarrow (1-\rho_0) \vlambda_0 + \rho_0\rnd{ \sqr{ \begin{array}{c} \hat{\alpha}_0-1 \\ N+\hat{\beta}_0-1 \end{array} } + \sum_{i=1}^N \mu_i \sqr{ \begin{array}{c} p_a(\vy_i)\\  p_b(\vy_i) \end{array} } } .
      \]
      where 
         $\hat{\alpha}_0 = \nabla_{\mu_0^{(1)}} \myexpect_q\sqr{-\half (\text{logit}(\pi_0) - m)^2}$ and 
         $\hat{\beta}_0 = \nabla_{\mu_0^{(2)}} \myexpect_q\sqr{-\half (\text{logit}(\pi_0) - m)^2}$ are the parameters of the pseudo beta prior. The gradients automatically convert the non-conjugate prior form into a conjugate one, which is due to \cite{khan2017conjugate}.
\end{ex}

The elegance of the update above is not a coincidence, rather it is by design. \cite{khan2017conjugate} linearize the whole VB objective with respect to all $\vmu$. This linearizes the non-conjugate terms while preserving the linearity of the conjugate terms; see \cite[Lemma 1]{khan2017conjugate}. This is a remarkable property of such linearizations. The BLR builds on this but it is the linearization which enables generalization to all sorts of algorithms. This insight is under-appreciated
so far, but we do hope that the examples shown in this paper motivates many readers to exploit such (Bayesian) linearization procedure.

\section{Future Work}

One drawback of the recipe is that it depends heavily on the $(\vlambda, \vmu)$ parameterization. There are several works that extend beyond this, for example, to mixture distributions \citep{linfast}, structured Gaussian distributions \citep{lin2021tractable}, and also to transformation families \citep{kiral2023lie}. Such approaches have shown to be useful in deriving existing algorithms, as well as designing new ones \citep{khan2021bayesian}. We believe that such extensions
exist for generic distributions and that it is possible to derive all sorts of updates in the same fashion as described in this paper.

\acks{I would like to thank Dr. Negar Safinianaini and other members of our group for discussions which not only compelled me to write this paper but also allowed me to see the value of techniques I have been using to derive VB for many years.}

\bibliography{refs}

\begin{thebibliography}{15}
\providecommand{\natexlab}[1]{#1}
\providecommand{\url}[1]{\texttt{#1}}
\expandafter\ifx\csname urlstyle\endcsname\relax
  \providecommand{\doi}[1]{doi: #1}\else
  \providecommand{\doi}{doi: \begingroup \urlstyle{rm}\Url}\fi

\bibitem[Bishop(2006)]{bishop2006pattern}
C.~M. Bishop.
\newblock \emph{Pattern Recognition and Machine Learning (Information Science
  and Statistics)}.
\newblock Springer-Verlag, Berlin, Heidelberg, 2006.
\newblock ISBN 0387310738.

\bibitem[Blei et~al.(2017)Blei, Kucukelbir, and McAuliffe]{blei2017variational}
D.~M. Blei, A.~Kucukelbir, and J.~D. McAuliffe.
\newblock Variational inference: A review for statisticians.
\newblock \emph{Journal of the American statistical Association}, 112\penalty0
  (518):\penalty0 859--877, 2017.

\bibitem[Hinton and Van~Camp(1993)]{hinton1993keeping}
G.~E. Hinton and D.~Van~Camp.
\newblock Keeping the neural networks simple by minimizing the description
  length of the weights.
\newblock In \emph{Annual Conference on Computational Learning Theory}, pages
  5--13, 1993.

\bibitem[Hoffman et~al.(2013)Hoffman, Blei, Wang, and
  Paisley]{hoffman2013stochastic}
M.~D. Hoffman, D.~M. Blei, C.~Wang, and J.~Paisley.
\newblock Stochastic variational inference.
\newblock \emph{The Journal of Machine Learning Research}, 14\penalty0
  (1):\penalty0 1303--1347, 2013.

\bibitem[Jaakkola and Jordan(1996)]{Jaakkola96b}
T.~Jaakkola and M.~Jordan.
\newblock A variational approach to {B}ayesian logistic regression problems and
  their extensions.
\newblock In \emph{International conference on Artificial Intelligence and
  Statistics}, 1996.

\bibitem[Khan and Lin(2017)]{khan2017conjugate}
M.~E. Khan and W.~Lin.
\newblock Conjugate-computation variational inference: converting variational
  inference in non-conjugate models to inferences in conjugate models.
\newblock In \emph{International Conference on Artificial Intelligence and
  Statistics}, pages 878--887, 2017.

\bibitem[Khan and Rue(2021)]{khan2021bayesian}
M.~E. Khan and H.~Rue.
\newblock The {B}ayesian learning rule.
\newblock \emph{arXiv preprint arXiv:2107.04562}, 2021.

\bibitem[Khan et~al.(2016)Khan, Babanezhad, Lin, Schmidt, and
  Sugiyama]{khan2016faster}
M.~E. Khan, R.~Babanezhad, W.~Lin, M.~Schmidt, and M.~Sugiyama.
\newblock Faster stochastic variational inference using proximal-gradient
  methods with general divergence functions.
\newblock In \emph{Proceedings of the Conference on Uncertainty in Artificial
  Intelligence}, 2016.

\bibitem[K{\i}ral et~al.(2023)K{\i}ral, M{\"o}llenhoff, and Khan]{kiral2023lie}
E.~M. K{\i}ral, T.~M{\"o}llenhoff, and M.~E. Khan.
\newblock The {L}ie-group {B}ayesian learning rule.
\newblock In \emph{International conference on Artificial Intelligence and
  Statistics}, 2023.

\bibitem[Lin et~al.(2019)Lin, Khan, and Schmidt]{linfast}
W.~Lin, M.~E. Khan, and M.~Schmidt.
\newblock Fast and simple natural-gradient variational inference with mixture
  of exponential-family approximations.
\newblock 2019.

\bibitem[Lin et~al.(2021)Lin, Nielsen, Khan, and Schmidt]{lin2021tractable}
W.~Lin, F.~Nielsen, M.~E. Khan, and M.~Schmidt.
\newblock Tractable structured natural-gradient descent using local
  parameterizations.
\newblock In \emph{International Conference on Machine Learning}, pages
  6680--6691. PMLR, 2021.

\bibitem[Paquet(2014)]{paquetconvergence}
U.~Paquet.
\newblock On the convergence of stochastic variational inference in {Bayesian}
  networks.
\newblock \emph{NIPS Workshop on variational inference}, 2014.

\bibitem[Saul et~al.(1996)Saul, Jaakkola, and Jordan]{saul1996mean}
L.~K. Saul, T.~Jaakkola, and M.~I. Jordan.
\newblock Mean field theory for sigmoid belief networks.
\newblock \emph{Journal of Artificial Intelligence Research}, 4:\penalty0
  61--76, 1996.

\bibitem[Tipping and Bishop(1999)]{tipping1999probabilistic}
M.~E. Tipping and C.~M. Bishop.
\newblock Probabilistic principal component analysis.
\newblock \emph{Journal of the Royal Statistical Society: Series B (Statistical
  Methodology)}, 61\penalty0 (3):\penalty0 611--622, 1999.

\bibitem[Winn and Bishop(2005)]{winn2005variational}
J.~Winn and C.~M. Bishop.
\newblock Variational message passing.
\newblock \emph{Journal of Machine Learning Research}, 6\penalty0
  (Apr):\penalty0 661--694, 2005.

\end{thebibliography}

\appendix

   \section{VB Stationary Point, The Linearity Property and Equivalence of \cref{eq:vb_fixed_pt,eq:blr_fixed_pt}}
   \label{app:linearity}

   {\bf VB stationary point:}
   We first give a short proof of the stationarity condition \cref{eq:vb_fixed_pt}, which essentially follows by rewriting the ELBO as a function of $q_i$ alone and expressing it as a Kullback-Leibler divergence (KLD) term,
   \begin{equation}
   \begin{split}
      \mathcal{L}(q) &= \myexpect_q [\log p(\vy,\vz)] + \sum_i \myexpect_q[ -\log q_i(\vz_i)] \\
      &= \myexpect_{q_i} \sqr{ \myexpect_{q_{/i}} [\log p(\vy,\vz)] -\log q_i(\vz_i) } + \text{const.}\\
      &= \dkls{}{q_i(\vz_i) }{\tilde{p}_i(\vz_i)} + \text{const.} 
   \end{split}
   \end{equation}
   where $\tilde{p}_i(\vz_i) \propto \exp[\myexpect_{q_{\backslash i}} [\log p(\vy,\vz)]]$ is a distribution where the rest $\vz_{\backslash i}$ is marginalized out. The minimum occurs when the two arguments in the divergence are equal, which gives us the condition in \cref{eq:vb_fixed_pt}.

   {\bf The linearity property:} This is due to the fact that the CE terms in log-joint are multi-linear in sufficient statistics \citep[p. 668]{winn2005variational}. Below, we give a formal statement which is the basis of Step 1.
\begin{thm}
   If the conditional $p(\vz_i|\vz_{\backslash i}, \vy)$ is conjugate to $q_i(\vz_i)$, that is, if there exists $\veta_i(\cdot)$ such that the log-conditional can be expressed in terms of $\vT_i(\vz_i)$ as follows,
   \begin{equation}
      \log p(\vz_i|\vz_{\backslash i}, \vy) = \myang{ \vT_i(\vz_i)\, ,\, \veta_i(\vz_{\backslash i}, \vy_i) } + \text{const.}, 
      \label{eq:logjoint}
   \end{equation}
   then $\myexpect_q[\log p(\vy, \vz)] = \myang{\vmu_i \, , \, \myexpect_{q_{\backslash i}} \sqr{ \veta_i(\vz_{\backslash i}, \vy_i) } } + \text{const.}$, which is linear with respect to $\vmu_i$.
\end{thm}
This also justifies Step 2 because the gradient with respect to $\vmu_i$ becomes
\begin{equation}
   \nabla_{\text{\vmeanparam}_i} \myexpect_q[\log p(\vy,\vz)] = \myexpect_{q_{\backslash i}} \sqr{ \veta_i(\vz_{\backslash i}, \vy_i) } 
   \label{eq:natgrad}
\end{equation}
which is simply the term in front of $\vmu_i$.

{\bf Equivalence:} Due to \cref{eq:blr_fixed_pt}, we can show that, at a fixed point, the optimal natural parameter is equal to the coefficient in front of $\vmu_i$,
\begin{equation}
   \vlambda_i^* = \nabla_{\text{\vmu}_i} \left. \myexpect_{q} \sqr{ \log p(\vy,\vz) } \right\vert_{\text{\vmu}_i = \text{\vmu}_i^*} = \myexpect_{q_{\backslash i}^*} \sqr{ \veta_i(\vz_{\backslash i}, \vy_i) },
   \label{eq:optlambda}
\end{equation}
   Using this, we can show that \cref{eq:blr_fixed_pt} leads to \cref{eq:vb_fixed_pt} by using the definition of $q_i^*(\vz_i)$, 
\begin{equation}
\begin{split}
   \log q_i^*(\vz_i) &= \myang{\vT(\vz_i), \vlambda_i^*} + c,\\ 
   &= \myang{\vT(\vz_i), \myexpect_{q_{\backslash i}^*} \sqr{ \veta_i(\vz_{/i}, \vy_i) } } + c\\
   &= \myexpect_{q_{\backslash i}^*} \sqr{ \log p(\vy, \vz) } + c', 
\end{split}
\end{equation}
   where the second and third line follow by using \cref{eq:optlambda} and \cref{eq:logjoint} respectively ($c$ and $c'$ are constants that do not depend on $\vz_i$). The derivation is reversible and we can also derive \cref{eq:blr_fixed_pt} from \cref{eq:vb_fixed_pt}, making the two statements equivalent whenever the posterior is a conjugate-exponential family.

{\bf Extension for non-constant base-measures:} If the exponential-family includes a base-measure $h_i(\vz_i)$ as shown below,
   \[q_i(\vz_i) \propto h_i(\vz_i) \exp\rnd{\myang{\text{\vT}_i(\text{\vz}_i), \text{\vnatparam}_i}},\]
   then the condition in \cref{eq:blr_fixed_pt} is modified to the following, 
   \begin{equation}
      \vlambda_i^* = \nabla_{\text{\vmu}_i} \left. \myexpect_{q} \sqr{ \log p(\vy,\vz) - \log h_i(\vz_i) } \right\vert_{\text{\vmu}_i = \text{\vmu}_i^*},
   \end{equation}
   The proof is exactly the same so we omit it.

\end{document}